\documentclass[sigconf]{acmart}

\usepackage{multirow}
\usepackage{makecell}

\AtBeginDocument{%
  }

\copyrightyear{2023}
\acmYear{2023}
\setcopyright{acmlicensed}\acmConference[MM '23]{Proceedings of the 31st ACM International Conference on Multimedia}{October 29-November 3, 2023}{Ottawa, ON, Canada}
\acmBooktitle{Proceedings of the 31st ACM International Conference on Multimedia (MM '23), October 29-November 3, 2023, Ottawa, ON, Canada}
\acmPrice{15.00}
\acmDOI{10.1145/3581783.3611961}
\acmISBN{979-8-4007-0108-5/23/10}

\begin{document}

\title{GridFormer: Towards Accurate Table Structure Recognition via Grid Prediction}


\author{Pengyuan Lyu}
\authornotemark[1]
\orcid{0000-0003-3153-8519}
\affiliation{%
  \institution{VIS, Baidu Inc.}
  \city{Shenzhen}
  \country{China}
}
\email{lvpyuan@gmail.com}

\author{Weihong Ma}
\authornote{Both authors contributed equally to this research.}
\affiliation{%
  \institution{VIS, Baidu Inc.}
  \city{Shenzhen}
  \country{China}
}
\email{scutmaweihong@gmail.com}
\orcid{0009-0007-2252-3856}

\author{Hongyi Wang}
\affiliation{%
  \institution{South China University of Technology}
  \city{Guangzhou}
  \country{China}
}
\email{eewanghy@mail.scut.edu.cn}
\orcid{0000-0002-5403-1449}

\author{Yuechen Yu}
\affiliation{%
  \institution{VIS, Baidu Inc.}
  \city{Shenzhen}
  \country{China}
}
\email{yuyuechen@baidu.com}
\orcid{0009-0006-9842-3360}

\author{Chengquan Zhang}
\authornote{Corresponding author.}
\affiliation{%
 \institution{VIS, Baidu Inc.}
  \city{Shenzhen}
  \country{China}
}
\email{zhangchengquan@baidu.com}
\orcid{0000-0001-8254-5773}

\author{Kun Yao}
\affiliation{%
 \institution{VIS, Baidu Inc.}
  \city{Beijing}
  \country{China}
}
\email{yaokun01@baidu.com}
\orcid{0000-0001-7155-4076}

\author{Yang Xue}
\affiliation{%
  \institution{South China University of Technology}
  \city{Guangzhou}
  \country{China}
}
\email{yxue@scut.edu.cn}
\orcid{0000-0002-1947-4957}

\author{Jingdong Wang}
\affiliation{%
 \institution{VIS, Baidu Inc.}
  \city{Beijing}
  \country{China}
}
\email{wangjingdong@baidu.com}
\orcid{0000-0002-4888-4445}

\renewcommand{\shortauthors}{Pengyuan Lyu et al.}

\begin{abstract}
All tables can be represented as grids. Based on this observation, we propose GridFormer, a novel approach for interpreting unconstrained table structures by predicting the vertex and edge of a grid. First,  we propose a flexible table representation in the form of an  $M \times N$ grid. In this representation, the vertexes and edges of the grid store the localization and adjacency information of the table. Then, we introduce a DETR-style table structure recognizer to efficiently predict this multi-objective information of the grid in a single shot. Specifically, given a set of learned row and column queries, the recognizer directly outputs the vertexes and edges information of the corresponding rows and columns. Extensive experiments on five challenging benchmarks which include wired, wireless, multi-merge-cell, oriented, and distorted tables demonstrate the competitive performance of our model over other methods.
\end{abstract}




\maketitle

\vspace{-1em}
\section{Introduction} 
Tables that contain rich structured information commonly appear in document images. Recently, with the rapid growth of demand for document AI~\cite{cui2021document,docvqa,structext-v2,layoutlm-v3}, table structure recognition, which plays the role of understanding tables from document images, has become significantly valuable.

Over the last few years, considerable research interests have been drawn to the field of table structure recognition. While tables with simple structures and clean backgrounds can be recognized well ~\cite{WANG20041479,liu2008identifying,ng1999learning,liu2009improving,kieninger1998table,doush2010detecting,itonori1993table,koci2018table,DeepDeSRT,GTE,LGPMA}, recognizing complicated table structures remains a challenging problem, which is primarily due to two main difficulties: 1) Firstly, tables in images vary widely in terms of structure and shape. For example, a table may be wired or wireless or a mixture of both of them. Additionally, the number of rows and columns can vary from just one to several hundred. 2) Secondly, some table images are taken from the wild, which suffer from complex backgrounds and geometrical distortions.

\begin{figure*}[t]
\centering
  \includegraphics[width=0.85\textwidth]{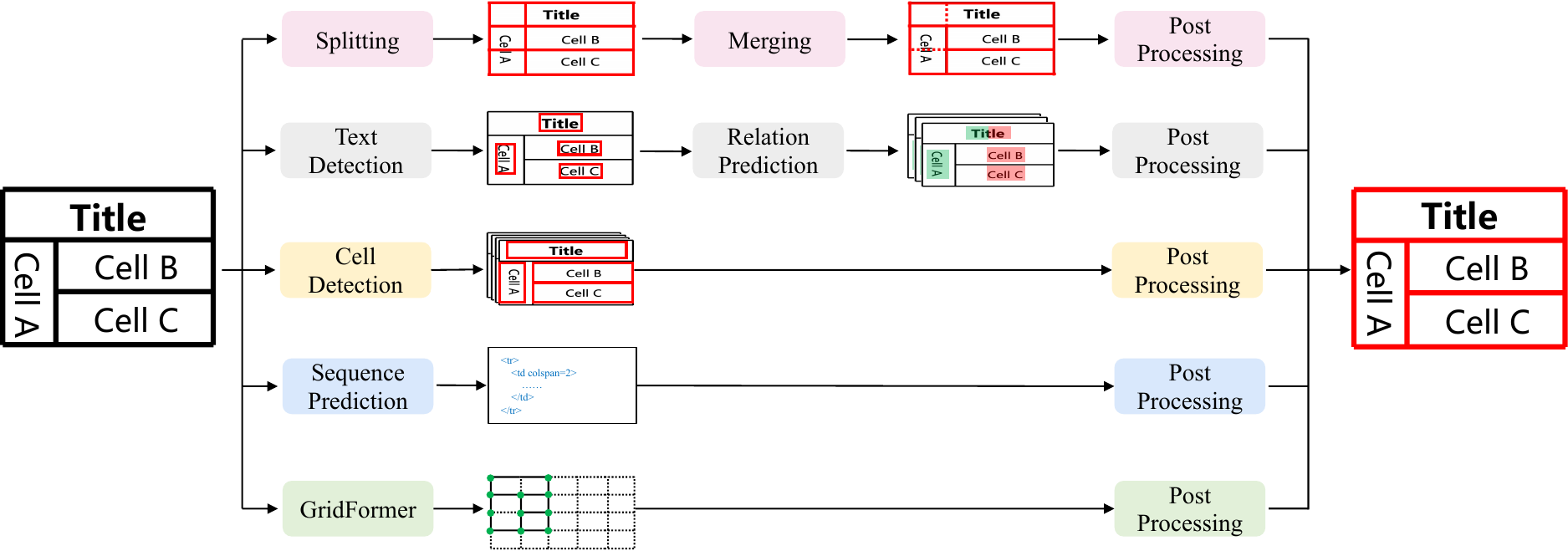} 
    \caption{Comparison between previous methods and ours. The pipelines are region-based methods, graph-based methods, cell-based methods, markup language-based methods and ours, respectively. The basic elements of the first four pipelines are over-split grid region, text instance, table cell and markup symbol respectively, while ours is the vertex and edge of the grid.} 
    \label{fig:introduction}
\vspace{-1.2em}
\end{figure*}

Recently, several methods have been proposed to tackle the challenge of recognizing table structures in complex scenarios. These methods can be generally categorized into four types, based on the table representation used: region-based, graph-based, cell-based, and markup language-based. For instance, region-based methods~\cite{DeepDeSRT,tabstruct,SEM,TSRFormer,trust} employ a split model to divide input table images into a grid of regions, and a merge model to combine over-split spanning cells. Graph-based research~\cite{DGCNN,FLAG,NCGM} treat detected text bounding boxes as table elements, construct graphs based on them, and use graph neural networks (GNNs) to predict whether two elements share a row or column, or cell. Alternatively, the cell-based method ~\cite{cycle_center_net} represents tables as cells, where CenterNet~\cite{centernet} is used to simultaneously predict the vertexes and center points of cells, and the resulting information is then used to reconstruct the overall table layout during the parsing process. In markup language-based studies~\cite{EDD,tablemaster,VAST}, a table image is expressed as a sequence of a markup language (e.g. HTML), and auto-regressive models are used to parse the table images. While previous methods have made significant progress in table structure recognition using various table representations, they still have noticeable limitations that affect their simplicity, flexibility, and effectiveness. For instance, 1) most of the existing methods are struggling with complex pipelines and low efficiency. 2) The majority of them are not all-rounders, making them less effective in handling complicated scenarios such as wireless tables~\cite{cycle_center_net}, oriented or distorted tables~\cite{tabstruct,tablenet,qasim2019rethinking,GTE,LGPMA}, tables with blank cells~\cite{EDD,tablemaster}, or tables without additional text annotations~\cite{DGCNN,FLAG,NCGM}.

 In this paper, we present GridFormer, a simple but effective table structure recognizer based on a flexible table representation. Our proposed representation is inspired by an observation that all tables can be represented by grids. Specifically, as shown in Figure~\ref{fig:representation}, the vertexes and boundaries of a cell in a table can be expressed by vertexes and edges in a grid, respectively.
 Vertexes on the grid that share the same logical indices (same row and column indices) as those on the table will be regarded as positive vertexes,
 and the corresponding physical coordinates will be inherited. In addition, edges belonging to the boundaries of the cells will be defined as positive edges, including the edges in the downward and rightward directions. With the positive vertexes and edges, a table can be restructured regardless of its appearance, shape, and structure. Note that, there are fundamental differences between the representation in ~\cite{DeepDeSRT,tabstruct,SEM,TSRFormer,trust} and ours despite the representation is also termed as "grid". In those methods, the "grid"  refers to regions that are divided by table splitting lines and require an extra merging module to combine them. In contrast, our representation is composed of vertexes and edges, which is more flexible and concise.

With the flexible grid representation, we propose an exceedingly simple yet highly effective model to recognize the table.
Specifically, we propose to use query selection modules to initialize the row and column reference points for the transformer decoder. 
Two parallel transformer decoders are adopted to decouple the predictions of rows and columns, allowing for more accurate restoration of table structure.
Inspired by DETR~\cite{DETR}, we employ three prediction heads to predict the class of rows and columns queries, the coordinates of vertexes, and the class of edges for reconstructing the input table.
Compared to the previous methods, our method enjoys two prominent advantages. 1) Our method is single-shot and end-to-end trainable, which greatly simplifies the complexity of table structure recognition pipeline. 2) Our method is robust to multiple complex scenarios, such as wired or wireless, oriented, and distorted tables. 

Extensive experiments are conducted to verify the effectiveness of our proposed method. 
With a simple pipeline, our method has achieved comparable or state-of-the-art performance on the public benchmarks, including PubTabNet~\cite{EDD}, FinTabNet~\cite{GTE}, SciTSR~\cite{graphTSR}, WTW~\cite{cycle_center_net},  and TAL~\cite{tal2021}. Besides, our method works pretty well on more challenging tables which are rotated and distorted.

The main contributions of this paper are summarized as follows:
\begin{itemize}
    \item 
    We present a flexible grid representation that enables restructuring a table regardless of its appearance, shape, or structure, using vertexes and edges information.
    \item 
    Our proposed approach is single-shot and end-to-end trainable, featuring a straightforward pipeline and robust capabilities. We leverage two parallel decoders to predict rows and columns information, respectively. Three prediction heads are employed to predict the vertex and edge information for the accurate table structure recognition output.
    \item Our method achieves satisfactory performance on multiple complicated table datasets. The results on multiple challenging scenarios show that our GridFormer is robust to unconstrained table images.
\end{itemize} 

\section{Related Work}

\subsection{Representations of Table}
Tables, as structured data, have undergone various representations in recent years, with four notable approaches emerging for table structure recognition. In the works of~\cite{DeepDeSRT,tabstruct,SEM,RobusTabNet,TSRFormer,trust}, a table is depicted as a grid of regions. These regions are obtained by dividing the table image using table lines, serving as the fundamental units of this representation. An additional merge model is required to combine these over-split regions. In ~\cite{DGCNN,GFTE,res2tim,FLAG,NCGM,e2eMMM}, table is treadted as a graph. The text instances within the table act as nodes, while the edges connecting the nodes predict their placement within rows, columns, or cells. In works such as ~\cite{GTE,li2021adaptive,cascadetabnet,LGPMA,cycle_center_net}, a table is represented by a group of cells. Some other information, such as center points and vertexes are also used to determine the adjacency relationship of cells. Differing from the aforementioned visual representations, ~\cite{deng2019challenges,EDD,tablemaster,table2latex} describe a table as a sequence of markup language (e.g., HTML and LaTeX) elements.

\subsection{Table Structure Recognizer}
\smallskip\noindent\textbf{Region-based methods.} 
In~\cite{split_merge,SEM,RobusTabNet,TSRFormer,trust}, a table is represented by a grid of regions. All of them follow a pipeline that splits the input table image into regions by row and column boundary segmentation and then merges the spanning cells based on features of adjacent regions. To achieve better performance when performing merging, SEM~\cite{SEM} incorporates both text and visual modalities to merge adjacent regions. To handle distorted tables, TSRFormer~\cite{TSRFormer} models the row or column boundaries as curves using several fixed-length points. With the continuous upgrading of the split and merge pipeline, more and more complicated scenarios can be handled. However, the two-stage framework is still a  limitation for these models to enjoy simple pipeline and end-to-end training, requiring further refinement.

\vspace{1mm}
\smallskip\noindent\textbf{Graph-based methods.} \cite{graphTSR,GFTE,qasim2019rethinking,TGRNet,NCGM,FLAG,e2eMMM} extract table elements (cells or text lines) first, then employ a graph network to learn the relation of the extracted table elements. GraphTSR \cite{graphTSR} applies graph-attention blocks to make the features of vertexes interact with the features of edges and eventually classify the adjacent vertexes in the horizontal and vertical directions respectively, deciding whether they are adjacent or not. TabStructNet \cite{tabstruct} combines table element detection and vertex relationship prediction into a single network, providing an end-to-end solution.  FLAG-Net \cite{FLAG} employs self-attention and graph networks to extract dense features and sparse features respectively, and designs gated networks to flexibly aggregate information. Besides, NCGM \cite{NCGM} enables the three modalities of geometry, appearance and content to collaborate with each other, and leverages modality interaction to boost the multi-modal representation for complex scenarios. The main limitation of those methods is that those methods rely on OCR annotations or results. However, the OCR annotations may not exist in the actual application scenario. And the performance of table structure recognition depends heavily on the precision of OCR, resulting in the potential error propagation from OCR to table structure recognition.

\vspace{1mm}
\smallskip\noindent\textbf{Cell-based methods.} 
~\cite{GTE,li2021adaptive,tabstruct, LGPMA,cycle_center_net} represent tables by a group of cells. GTE \cite{GTE} uses an object detection-based method to detect cells directly and uses heuristic rules in post-processing to recover the table structure.  LGPMA \cite{LGPMA} applies soft pyramid masks at the local and global levels, allowing the model to detect cell boundaries of wireless tables more accurately. Cycle-CenterNet \cite{cycle_center_net} goes a step further by refining the detection granularity to the cell vertexes. The network uses two branches to detect cell center points and vertexes and uses the cycle-pairing module to determine the attribution of vertexes. Since the neighboring cells are co-vertexes, the table structure can be obtained using the heuristic rules. These methods work well on the wired table but may struggle with the wireless table which is not friendly to conduct cell detection.

\vspace{1mm}
\smallskip\noindent\textbf{Markup language-based methods.}  Since table structure can be represented by a sequence of a markup language, some methods \cite{deng2019challenges,EDD,table2latex,tablemaster,VAST} treat the task of table recognition as an image-to-sequence translation task and apply encoder-decoder models to transform the input table image. The encoder acquires the visual features of the table image and two independent structure decoders are employed to output the table structure and cell positions respectively. Such methods rely on large amounts of data for training and suffer from cumulative errors, attentional bias, and low efficiency, which is inefficient and may not work well on large tables (they have long sequences).

\section{GridFormer}
In this section, we first introduce our proposed grid representation for the arbitrary table. After that, we describe the details of the network, training process, and inference.

\begin{figure}[t]
\centering

    \includegraphics[width=0.45\textwidth]
    {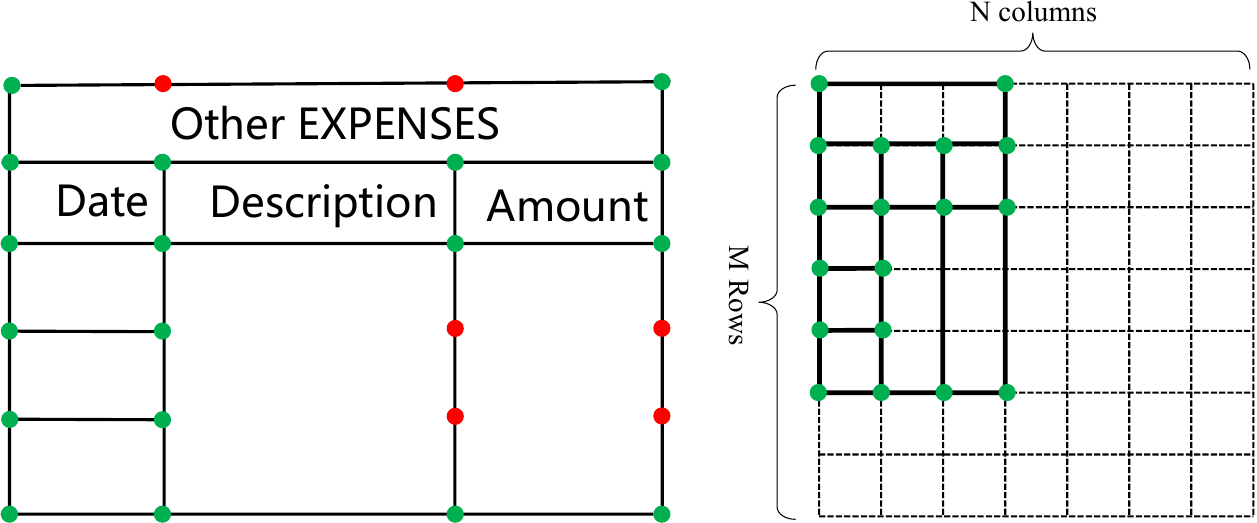} 
    \caption{Illustration of the proposed grid representation, (a) and (b) are table images and the corresponding grid representation. A table with an arbitrary structure can be represented by a $M \times N$ grid. The vertexes in green are positive vertexes and the red vertexes are negative. The Solid and dotted lines denote the positive and negative edges respectively.}
    \label{fig:representation}
\vspace{-1.2em}
\end{figure}

\begin{figure*}[t]
\centering
  \includegraphics[width=0.85\textwidth]{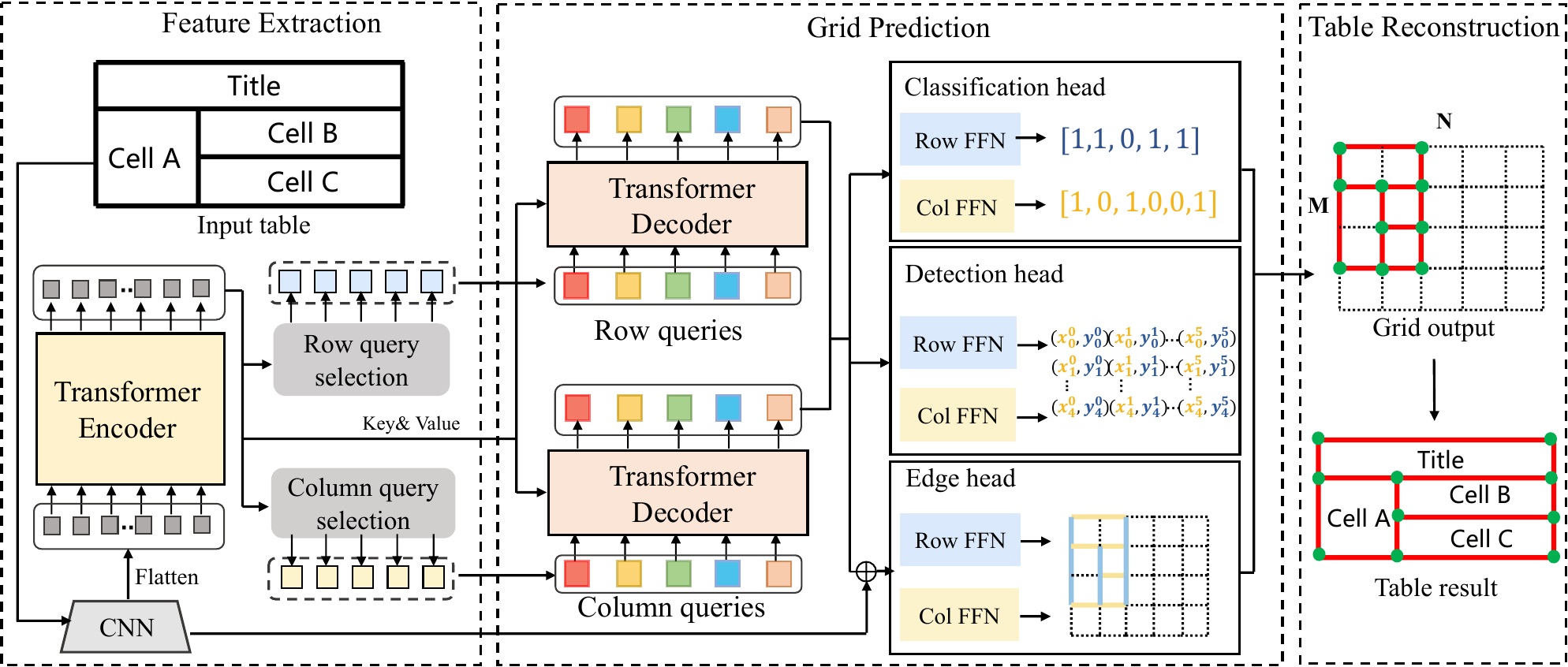} 
    \caption{Illustration of the proposed GridFormer. The feature extraction module obtains the feature representation and generates the row and column reference points that are fed into the transformer decoder. Given a set of row queries and column queries, we predict the class, vertexes, and edges of the corresponding rows and columns.}
    \label{fig:pipeline}
\vspace{-1.5em}
\end{figure*}

\subsection{Grid Representation}
Tables can vary in their number of rows, columns, and structures. However, all tables share a common characteristic that they can be normalized to a fixed-size grid.
Based on this observation, we propose a new table representation that represents a table as a grid. As shown in Figure~\ref{fig:representation}, given a table image, we can transform it into a grid easily.
Specifically, we utilize the grid vertexes to represent the vertexes of the table. The vertexes having the same relative position (same row and column indices) as the vertexes of cells are considered positive, while the remaining vertexes on the grid are regarded as negative. Additionally, we employ the edges of the grid to represent the cell boundaries. 
If an edge is part of a cell boundary, it is considered positive, while edges that are not part of any cell boundary are considered negative. Theoretically, the structure of a table can be reconstructed with the positive vertexes and edges. To further restore the physical coordinates of a table, we also store the coordinates information of each positive vertex. 

Another design of our grid representation is that, for a given set of table images, 
all grid representations will be padded to the same shape. By default, the padding value is set to negative.
In other words, the table images of a given set will be represented by grids that share a fixed  $M \times N$ size. And the $M$ and $N$ are always greater than the largest rows and columns of the table set. This design brings a potential advantage. In detail, the table images set can be transformed into a grid with a fixed-length sequence, consisting of vertexes and edges in order. And the standardized representation will greatly simplify the network design of grid prediction, which will be described in section~\ref{section:network}.

\subsection{Network}
\label{section:network}
With the new proposed grid representation for a table, we can restructure a table via grid prediction. To achieve this goal, a single-shot and end-to-end trainable recognizer is proposed. The recognizer is adapted from Deformable DETR~\cite{deformable_detr} and consists of a feature extraction module and a grid prediction module. As shown in Figure~\ref{fig:pipeline}, given an input table image, the feature extraction module extracts a compact feature representation, generates reference points for rows and columns, and the grid prediction module directly predicts the vertexes and edges to restore the table structure.

\subsubsection{Feature Extraction}
We utilize a CNN as our backbone model to extract image features. The transformer encoder which has the ability to capture long dependency of features is also employed to enhance the image representation.
Inspired by ~\cite{deformable_detr}, we propose query selection module to generate reference points for rows and columns. These reference points are then utilized as initial reference points for the transformer decoder.

\smallskip\noindent\textbf{Backbone.}
Following ~\cite{DETR}, we use ResNet-50\cite{resnet} as our backbone. Mathematically, input an image  $\mathsf{I} \in \mathbb{R}^{3\times H \times W}$, the backbone generates multi-scale feature maps $\left \{ x^{l}  \right \}_{l=1}^{5}$, corresponding to the output of the 5 stages of ResNet. Where $\mathsf{x} \in \mathbb{R}^{C^{l}\times H^{l} \times W^{l}}$. $H^{l}, W^{l}$ are the shape of $x^{l}$, which is of resolution $2^{l}$ lower than the input image.

\smallskip\noindent\textbf{Transformer encoder.} We use deformable transformer encoder ~\cite{deformable_detr} to capture the long dependency of features. The multi-scale deformation attention module fuses the features of multi-scale and decreases the complexity. In detail, the features from stage 3 to stage 5 are used. We first use a $1 \times 1$ convolution to transform all the multi-scale features to the channel of 256. After that, 6 stacked deformable transformer encoder layers are employed in default, and the richer feature maps $\left \{ f^{l}  \right \}_{l=3}^{5}$  are obtained.

\smallskip\noindent\textbf{Query selection.} 
We employ the query selection module output to initialize reference points for the transformer decoder. During the training phase, we add supervision on the query selection module. The labels for each row reference point are set to the y-mean value of the corresponding row, while the labels for each column reference point are set to the x-mean value of the corresponding column. Different from ~\cite{deformable_detr}, which generates proposals at each pixel on the feature map, we generate point proposals at fixed position $\tau$. The row proposals are evenly located at position $(\tau_{1}, y_{i})$, and the column proposals are evenly located at position $(x_{i}, \tau_{2})$, where $\tau_{1}=W/4$, $\tau_{2}=H/4$, $x_{i}\in[0, W^{l}]$, $y_{i}\in[0, H^{l}]$ with step of 1. Here, $H^{l}, W^{l}$ are the dimension of the encoded feature map. The query selection module outputs the positive probability and the regression result of reference points. The top-80 scoring proposals are picked directly and no NMS is applied before feeding the reference points to the grid prediction module.

\subsubsection{Grid Prediction}
\label{subsub:grid_pred}
The grid prediction module consists of a two-stream decoder with three task-specific FFN heads. These two decoders are responsible for the predictions of the vertexes and edges in the row direction and the column direction of the table.

\begin{figure}[t]
\setlength{\abovecaptionskip}{1em}
\centering
 \includegraphics[width=0.45\textwidth]{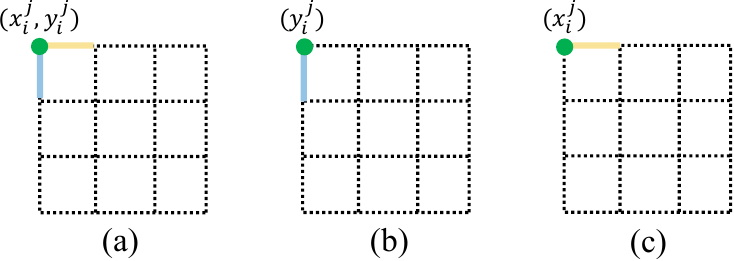} 
    \caption{The necessary predictions of different settings to restructure a table. (a) is the case of only using one decoder to predict all rows or all columns. (b) and (c) are the decoupled version of (a), which use row queries and column queries to predict all rows and columns, respectively.} 
    \label{fig:decoder}
\vspace{-2.1em}
\end{figure}

\smallskip\noindent\textbf{Two-stream decoder.} To recover the table structure accurately, we require information about the grid, including positive rows and columns, their physical coordinates on the x-axis and y-axis, and the edges connecting the subsequent vertexes in the right and down directions. Theoretically, it is possible to utilize a single decoder to restore a grid by predicting the vertexes and edges information of all rows or columns simultaneously. Nevertheless, this design may incur a performance penalty compared to the decoupled decoder. To illustrate this, let's consider the position prediction using a single decoder as an example. Typically, the range of coordinates along the y-axis for vertexes within the same row is not significantly diverse, while the coordinates along the x-axis can exhibit substantial variation, making accurate localization challenging.

To mitigate the above-mentioned issue, we propose to decouple the predictions of rows and columns, allowing for a more accurate restoration of the table structure. Specifically, we use two parallel transformer decoders to transform the learned row queries $\mathsf{Q}_{row}$ and column queries $\mathsf{Q}_{col}$ to row embeddings $\mathsf{Z}_{row}$ and column embeddings $\mathsf{Z}_{col}$ respectively. As shown in Figure~\ref{fig:decoder}, we decouple the prediction of (a) to (b) and (c), which corresponds to the row predictions and column predictions, respectively. In this way, the row embedding predicts only the coordinates on the y-axis and edges in the down direction, while the column embedding predicts only the coordinates on the x-axis and edges in the right direction.

We follow Deformable DETR~\cite{deformable_detr} and employ 6 deformable transformer decoder layers to obtain output embeddings. The row decoder and column decoder take row queries $\mathsf{Q}_{row}$ and column queries $\mathsf{Q}_{col}$ as input respectively.
The input queries conduct cross attention to a small set of key sampling points around reference points on the extracted feature maps of the input image.
After that, the row embeddings $\mathsf{Z}_{row} = \left \{ z_{row}^{l}  \right \}_{l=1}^{6}$ and column embeddings $\mathsf{Z}_{col} = \left \{ z_{col}^{l}  \right \}_{l=1}^{6}$ that contain global information of rows and columns are yielded, where $z_{row}^{l} \in \mathbb{R}^{M \times d}$, $z_{col}^{l} \in \mathbb{R}^{N \times d}$.

\smallskip\noindent\textbf{Prediction heads.}
We follow DETR~\cite{DETR} and use FFNs which consist of a 3-layer perceptron with ReLU activation function to compute the final prediction. Three prediction heads are employed to predict the class of rows and columns, the coordinates of vertexes, and the class of edges for reconstructing the input table.

1) Row and column classification: Given row embedding $z_{row}^{l} \in \mathbb{R}^{M \times d} $ and column embedding $z_{col}^{l} \in \mathbb{R}^{N \times d}$ , the FFNs in the classification head predict the probability $\mathsf{\hat{p}}_{row}^{l} \in \mathbb{R}^{M}$ of rows and the $\mathsf{\hat{p}}_{col}^{l} \in \mathbb{R}^{N}$ of columns.

2) Position regression: There is a stronger correlation between y-axis/x-axis values within the same row/column. Therefore, we use the FFNs to predict normalized coordinates within the same row/column from the row/column embedding. The row FFNs in the detection head predict the normalized y-axis coordinates $\mathsf{\hat{r}}_{row}^{l} \in \mathbb{R}^{M \times N}$ of vertexes, while the column FFNs predict the normalized x-axis coordinates $\mathsf{\hat{r}}_{col}^{l} \in \mathbb{R}^{N \times M}$ of vertexes.

3) Edge classification: Different from the position of vertexes, edges in the downward or rightward direction are less relevant within the same row/column. Therefore, we extract useful features from vertexes for edge binary classification by considering both global and local information. Concretely, we first use the matched indices result from the Hungarian algorithm to re-order row and column queries predictions on y-axis/x-axis, creating the $M \times N$ grid. We can then obtain the coordinates of each vertex $\mathbb{V} \in \mathbb{R}^{M \times N}$ in the grid by using the predicted normalized y-axis values $\mathsf{\hat{r}}_{row}^{l}$ and x-axis values $\mathsf{\hat{r}}_{col}^{l}$.
To incorporate global information, we repeat the row embedding for vertexes on the same row and the column embedding for vertexes on the same column. For local information, we use position prediction to sample visual features on CNN visual backbone. The concatenated feature of row embeddings and visual feature is used to predict the edges $\mathsf{\hat{e}}_{row}^{l} \in \mathbb{R}^{N \times M}$ that are in the downward, while the edges in the rightward $\mathsf{\hat{e}}_{col}^{l} \in \mathbb{R}^{N \times M}$ are predicted by the concatenated feature of column embeddings and visual feature.

\subsection{Loss Function}
\label{subsection:loss}
Since the row/column query output is un-ordered, following~\cite{DETR}, we use bipartite matching to assign the ground truth to predictions. We compute the matching cost with the class prediction and the $L1$ distance between prediction reference points and ground truth of rows and columns. The match indices are computed using the Hungarian algorithm. After that, the label of row/column classification, position regression, and edge classification  can be obtained.

\smallskip\noindent\textbf{Row and column classification.}
We use Focal loss~\cite{focal} to optimize the task of row and column classification. The loss function is formulated as follows,

\begin{equation}
    L_{cls} = Focal(\mathsf{\hat{p}}_{row},  \mathsf{p}_{row}) + Focal(\mathsf{\hat{p}}_{col},  \mathsf{p}_{col}).
\end{equation}

\smallskip\noindent\textbf{Position regression.} We optimize the regression of vertexes with two objective functions. First, the most commonly-used L1 loss is applied to all positive vertexes. In addition, we also design a global loss that guides learning from the perspective of the cell level. We compute the bounding rectangles of each cell from the original predictions and labels of vertexes and define them as $\mathsf{\hat{g}} \in \mathbb{R}^{K \times 4}$ and $\mathsf{g} \in \mathbb{R}^{K \times 4}$, where $K$ is the number of cells in the input table image. We use the generalized IoU loss~\cite{giou} to minimize the difference between $\mathsf{\hat{g}}$ and $\mathsf{g}$. Overall, the loss of vertex position regression is defined as 

\begin{equation}
\begin{aligned}
    L_{coord} = L1(\mathsf{\hat{r}}_{row},  \mathsf{r}_{row}) + L1(\mathsf{\hat{r}}_{col},  \mathsf{r}_{col}) + \gamma_{1}L_{iou}(\mathsf{\hat{g}}, \mathsf{g}).
\end{aligned}
\end{equation}

Besides, reference point supervision is also added on the query selection module and the decoder stage. L1 loss is used to optimize the learning of reference points, which is formulated as 
\begin{equation}
    L_{ref} = L1(\mathsf{\hat{m}}_{row},  \mathsf{m}_{row}) + L1(\mathsf{\hat{m}}_{col},  \mathsf{m}_{col}).
\end{equation} Note that the loss is only applied to the positive rows and columns.

\smallskip\noindent\textbf{Edge classification.} The focal loss is also used to optimize the task of edge prediction to mitigate the extreme imbalance between the positive and negative edges. We formulate the objective function as follows:
\begin{equation}
    L_{edge} = Focal(\mathsf{\hat{e}}_{row},  \mathsf{e}_{row}) + Focal(\mathsf{\hat{e}}_{col},  \mathsf{e}_{col}).
\end{equation}

The whole objective function is a combination of the above-mentioned losses, which is given as 
\begin{equation}
    L = \lambda_{1}L_{cls} + \lambda_{2}L_{coord} + \lambda_{3}L_{ref} + \lambda_{4}L_{edge}.  
\end{equation}

We set the $\lambda_{1}$, $\lambda_{2}$, $\lambda_{3}$, $\lambda_{4}$, and $\gamma_{1}$ to 1,5,5,5 and 0.1 respectively to balance the different tasks. Besides, following ~\cite{DETR}, the auxiliary decoding losses of each decoder layer are also used. Note that the model is trained in an end-to-end manner with multi-tasks, which is also the strength of our proposed method.

\subsection{Table Reconstruction}
With the predicted grid information, we can restructure a table easily. In detail, we first obtain the positive rows and columns via a threshold $\tau_{1}$.
The prediction whose score is higher than $\tau_{1}$ is thought as positive row/column.
We use the reference point predictions from row queries to sort the row predictions on the y-axis. Similarly, the column predictions are sorted on the x-axis by the reference point predictions from column queries.
After that, we use another threshold $\tau_{2}$ to get all positive edges. Edges with a score greater than $\tau_{2}$ are considered positive edges. Next, the positive vertexes can be classified with the help of positive edges. Theoretically, a vertex must be a positive vertex if there are more than 2 positive edges connected to it, except the four corner points of the grid. 

Based on the positive vertexes and edges, the structure of a table can be built, and the physical coordinate of each positive vertex can be obtained in $\mathsf{\hat{r}}_{row}^{l}$ and $\mathsf{\hat{r}}_{col}^{l}$. Finally, all cells are yielded by executing the breadth-first algorithm on the grid to group the adjacent positive vertexes.

\section{Experiments}
\subsection{Datasets}

We conducted a comprehensive validation of our model's performance on various datasets, including well-known regular table benchmarks such as SciTSR, PubTabNet, and FinTabNet, which are derived from PDF documents. Additionally, we evaluated our model on scene table benchmarks like WTW and TAL, which consist of tables collected from real-life scenarios. The regular benchmarks encompass a wide range of table structures, including long tables, wired or wireless tables, and multi-merge-cell tables. This diverse set of structures poses a significant challenge in effectively restructuring and accommodating the various structural variations encountered. Furthermore, the scene table benchmarks comprise tables with curved, oriented, and distorted structures, which further intensifies the difficulty in handling various distortions.

\smallskip\noindent\textbf{SciTSR}~\cite{graphTSR} is a dedicated dataset created to address the task of table structure recognition in scientific papers. The dataset consists of 12,000 training samples and 3,000 test samples, all available in PDF format. In line with previous work, we evaluate the performance of our model using the cell adjacency relationship metric~\cite{table2013}.

\smallskip\noindent\textbf{PubTabNet}~\cite{EDD} is a large-scale dataset containing 500,777 training images, 9,115 validation images, and 9,138 testing images. The tables in this dataset are extracted from scientific documents and exhibit complex structures, including wireless cells, spanning cells, empty cells, and variations in the number of rows and columns. It serves as a popular benchmark for evaluating the robustness of table recognition models in handling tables with intricate structures. Due to the absence of released annotations for the test set, we follow previous approaches~\cite{LGPMA,SEM,GTE,TSRFormer} and evaluate our model on the validation set using TEDS and TEDS-Struct~\cite{EDD} metrics.

\smallskip\noindent\textbf{FinTabNet}~\cite{GTE} is a large dataset containing financial tables. The tables in this dataset have fewer graphical lines and larger gaps than those found in tables from scientific documents, and they exhibit more color variations. The dataset is split into 92k training images, 10,635 validating images and 10,656 testing images. Following ~\cite{GTE, tableformer}, we use the split set of training for training and validating samples for testing. We also use TEDS-Struct~\cite{EDD} to evaluate table structure recognition performance.

\smallskip\noindent\textbf{WTW}~\cite{cycle_center_net} images are collected in the wild. The non-rigid image deformation and oriented tables with complicated image background pose a great challenge for accurate table structure recognition. The dataset is split into training/testing subsets with 10,970 and 3,611 samples respectively. Following ~\cite{TSRFormer}, we crop table regions from original images for both training and testing. Following ~\cite{cycle_center_net, TSRFormer}, we use cell adjacency relationship as the evaluation metric.

\smallskip\noindent\textbf{TAL\_OCR\_TABLE}~\cite{tal2021} is also large-scale and high-quality, which is used in table structure recognition competitions. The images in this dataset are all taken from the wild, which have a complex background and geometric distortion. Due to the annotations of test set is not released, we randomly divided the original training set into a new training set and test set, with the numbers 12285 and 3000, respectively. The split set of filenames will be released. We name this dataset TAL for simplicity.
To further explore the robustness of a table structure recognizer on more challenging scenarios, we also propose TAL\_rotated and TAL\_curved based on TAL. TAL\_rotated contains rotated images with angles selected randomly from $-30^{\circ}$ to $30^{\circ}$, while TAL\_curved is a dataset with geometrically distorted images. We have provided image examples in the supplemental material. We use TEDS-struct~\cite{EDD} to evaluate the performance of structure recognition. The F-Score of the predicted cells and the ground truth with an IOU of 0.6 is also used to evaluate the localization performance.

\begin{table*}[t]
\setlength{\abovecaptionskip}{0em}
\caption{Results of logical structure recognition and physical coordinates prediction on TAL, TAL\_rotated, TAL\_curved.}  
\centering
\setlength{\tabcolsep}{12pt}
\begin{tabular}{lcccccc}
\hline
 \multirow{2}*{Method} & \multicolumn{2}{c}{TAL}&\multicolumn{2}{c}{TAL\_rotated}&\multicolumn{2}{c}{TAL\_curved} \\
 \cline{2-7}
 & TEDS-Struct &  F-Score & TEDS-Struct & F-Score & TEDS-Struct & F-Score \\
\hline
  SPLERGE~\cite{split_merge} &91.5  &58.3 &74.9&3.84 &63.0 &14.6 \\ 
  TableMaster~\cite{tablemaster}&98.8  &80.8 &98.2&27.8 &98.5 &64.5\\ 
  \hline
  GridFormer &\textbf{99.4} &\textbf{98.9}&\textbf{99.1} &\textbf{92.9} &\textbf{99.2} &\textbf{96.8}\\ 
\hline
\end{tabular}
\label{tab:tal}
\end{table*}

\begin{table}[h]
\setlength{\abovecaptionskip}{0em}
 \centering
	\caption{Comparision on PubTabnet and FinTabNet. FT: Model was trained on PubTabNet then finetuned.}
	\begin{tabular*}{\hsize}{@{\extracolsep{\fill}}lcc}
		\toprule
	    \multicolumn{3}{c}{\textbf{PubTabNet}} \\
		\midrule
		Methods       & TEDS &  TEDS-Struct  \\
		\midrule
            EDD~\cite{EDD}                  &88.3  & -   \\
            TabStruct-Net~\cite{tabstruct}  &-     & 90.1  \\
            GTE~\cite{GTE}                  &-  & 93.0  \\
            SEM~\cite{SEM}                  &93.7  & -  \\
            LGPMA~\cite{LGPMA}              &94.6  & 96.7   \\
            FLAG-Net~\cite{FLAG}            &95.1  & -    \\
            NCGM~\cite{NCGM}                &95.4  & -   \\
            TableFormer~\cite{tableformer}  &93.60 & 96.75   \\
            TSRFormer ~\cite{TSRFormer}     &-   & \textbf{97.5} \\
            TRUST ~\cite{trust}             &96.20   & 97.10 \\
            VAST ~\cite{VAST}               &\textbf{96.31} & 97.23 \\
            \textbf{GridFormer}      & 95.84  & 97.0 \\
		\bottomrule
            \multicolumn{3}{c}{\textbf{FinTabNet}} \\
		\midrule
            EDD~\cite{EDD}  & - & 90.6 \\
            GTE~\cite{GTE}  & - & 87.1 \\
            GTE(FT)~\cite{GTE}  & - & 91.0 \\
            TableFormer~\cite{tableformer} & - & 96.8 \\
            VAST ~\cite{VAST}   & - & \textbf{98.63} \\
            \textbf{GridFormer} & - & \textbf{98.63} \\
            \bottomrule
	\end{tabular*}
	\label{tab:fintabnet}
\end{table}

\begin{table}[h]
\setlength{\abovecaptionskip}{0em}
    \centering
    \caption{Comparison of cell adjacency relation score on the SciTSR. * denotes the evaluation results without taking empty cells into account.}
    \begin{tabular*}{\hsize}{@{\extracolsep{\fill}}lcccc}
        \toprule
        Methods       & Training Dataset & P~(\%)& R~(\%) & F1~(\%)  \\
        \midrule
        GraphTSR\cite{graphTSR}      & SciTSR  & 95.90  & 94.80  & 95.30             \\
        TabStructNet\cite{tabstruct} & SciTSR  & 92.70  & 91.30  & 92.00             \\
        LGPMA\cite{LGPMA}            & SciTSR  & 98.20  & 99.30  & 98.80     \\
        SEM\cite{SEM}                & SciTSR  & 97.70  & 96.52  & 97.11     \\
        RobustTabNet\cite{RobusTabNet} & SciTSR & 99.40 & 99.10  & 99.30 \\
        FLAG-Net\cite{FLAG}            & SciTSR & 99.70 & 99.30  & 99.50 \\
        VAST \cite{VAST}               & SciTSR & 99.77 & 99.26  & 99.51 \\ 
        NCGM\cite{NCGM}                & SciTSR & \textbf{99.70} & \textbf{99.60} & \textbf{99.60} \\
        TSRFormer\cite{TSRFormer}      & SciTSR & \textbf{99.70} & \textbf{99.60} & \textbf{99.60} \\
        \hline
        \textbf{GridFormer}          & SciTSR   & 99.36 & 99.04 & 99.20 \\
        \textbf{GridFormer*}         & SciTSR   & 99.46 & 99.14 & 99.30 \\
        \bottomrule
    \end{tabular*}
    \label{tab:sciTSR}
\end{table}

\subsection{Implementation Details}
We use ResNet-50 as the backbone, followed by 6 deformable transformer encoder layers and 6 deformable transformable decoder layers. We find the performance is less sensitive to the value of row query number and column query number.
Hence, we match the number of row/column queries to the largest number of rows and columns in the corresponding dataset for computational efficiency. We train the networks end-to-end with the AdamW optimizer and set an initial learning rate of 2e-4.
All models are trained on 8 V100 GPUs with a total batch size of 24. We use the multi-scale training strategy to train all models. The short side of the input image is scaled to a value randomly selected from a list of [384, 416, 448, 480, 512]. We keep the aspect ratio of an input image and limit the long side to no large than 640.
In the inference stage, we resize the long side of the input image to 640 and keep the aspect ratio. The score threshold $\tau_{1}$ and $\tau_{2}$ are set to 0.5 and 0.4 respectively for all experiments.

\subsection{Comparasion with previous state-of-the-arts}

\begin{figure*}[t]
\centering
  \includegraphics[width=0.85\textwidth]{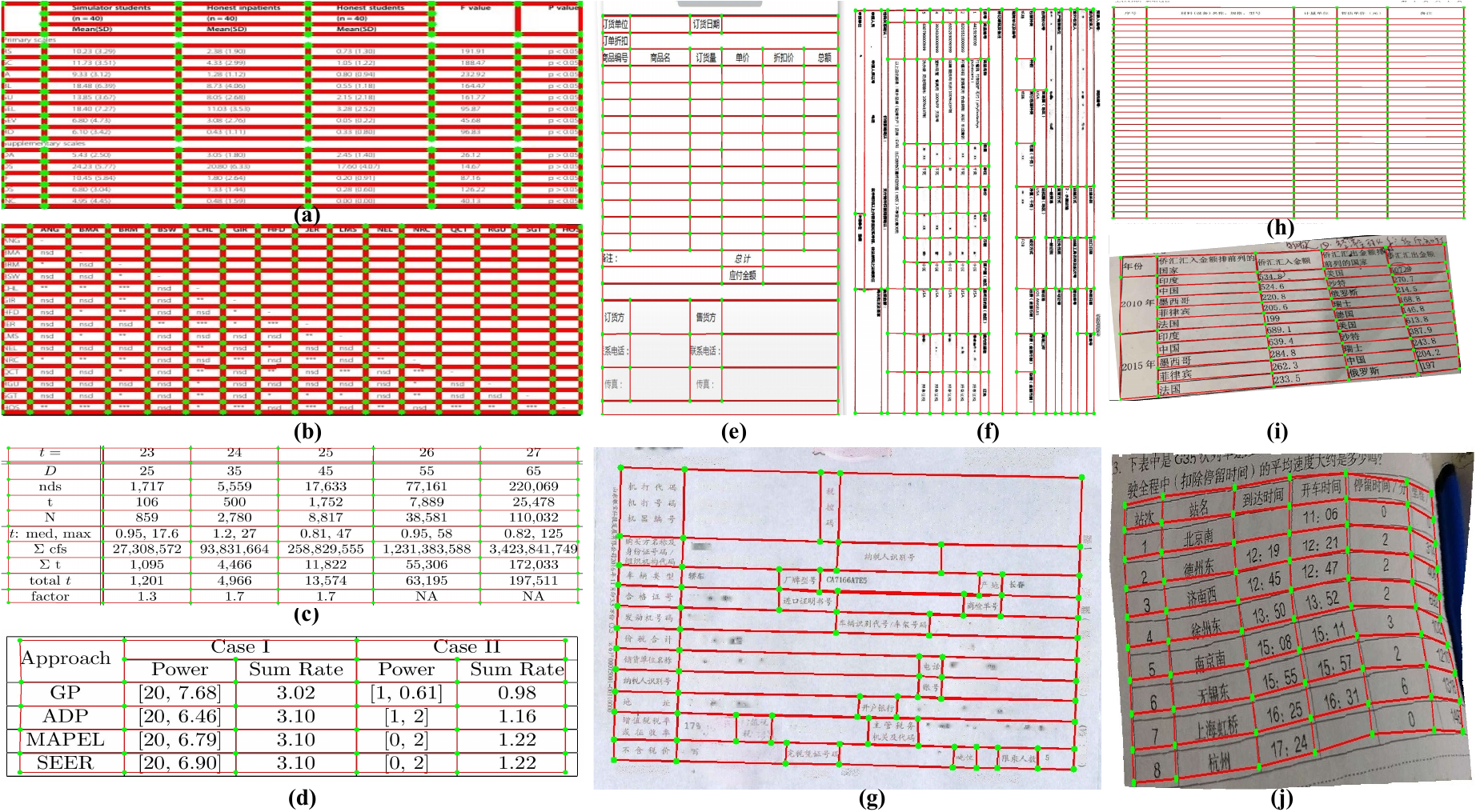} 
    \caption{Prediction results from GridFormer. (a-b) are from PubTabNet, (c-d) are from SciTSR, (e-g) are from WTW, (h-j) are from TAL, TAL\_rotated, and TAL\_curved. We visualize the edges with red lines and the positive vertexes with green circles.}
    \label{fig:vis_prediction}
\end{figure*}

\smallskip\noindent\textbf{Performance on regular tables.}
We evaluated our proposed method on regular tables that were scanned from PDF documents, and compared it with several state-of-the-art methods on PubTabNet, FinTabNet, and SciTSR datasets, and the results are reported in Table~\ref{tab:fintabnet} and Table~\ref{tab:sciTSR}. For PubTabNet, we report the TEDS and TEDS-Struct simultaneously. It is noted that the OCR results of PubtabNet are obtained from the text detection method PSENet~\cite{psenet} and text recognition method MASTER~\cite{master}. We linked each cell with the corresponding text results following \cite{tablemaster}. Our method achieves 95.84\% on TEDS and 97.0\% on TEDS-Struct, which is comparable to the previous method TSRFormer and outperforms other methods. On FinTabNet, our method achieves a TEDS-Struct score of 98.6\%, improving the score by 1.8\% compared to the competitive method TableFormer.
On the SciTSR dataset, where most methods reach 99\% performance, our proposed method achieves an F1-score of 99.3\% when ignoring empty cells, which is comparable to other methods. These results demonstrate the excellent performance of our proposed method on regular tables from scanned documents. As shown in Figure~\ref{fig:vis_prediction}, we visualize the performance of this method when facing long tables, wired or wireless tables, and multi-merge-cell tables.

\smallskip\noindent\textbf{Performance on complex tables.}
To evaluate the performance of our proposed method on real-scene tabular images, we conduct comparisons with competitive methods on the publicly available WTW and TAL datasets. Table~\ref{tab:wtw} presents the results on the WTW dataset, where GridFormer achieves the highest F1-score of 94.1\%, outperforming the second-best method by 0.7\%, which is on par with NCGM~\cite{NCGM}.
Furthermore, to assess the effectiveness of our method in oriented and distorted scenarios, we compare it with two mainstream open-source methods, SPLERGE~\cite{split_merge} and TableMaster~\cite{tablemaster}, on the TAL, TAL\_rotated, and TAL\_curved datasets. As shown in Table~\ref{tab:tal}, our method achieves TEDS-Struct scores of 99.4\%, 99.1\%, and 99.2\% on the original, rotated, and curved datasets, respectively. Notably, the performance gaps between our method and other competitors in terms of localization are substantial, with our method outperforming the second-best method by 18.1\% to 65.1\%. These results demonstrate the accurate regression of table vertexes by our method, even in complex scenarios, which can also be viewed in Figure~\ref{fig:vis_prediction}.

\begin{table}[t]
    \centering
    \setlength{\abovecaptionskip}{0em}
    \caption{Results on WTW dataset.}
    \begin{tabular*}{\hsize}{@{\extracolsep{\fill}}ccccc}
    \toprule
     Methods       &  Prec.(\%)& Rec.(\%) & F1-score(\%)  \\
    \midrule
    Cycle-CenterNet~\cite{cycle_center_net} & 93.3 & 91.5 & 92.4 \\
    TSRFormer~\cite{TSRFormer}              & 93.7 & 93.2 & 93.4 \\
    NCGM\cite{NCGM}                         & 93.7 & \textbf{94.6} & \textbf{94.1} \\
    \hline
    \textbf{GridFormer}                     & \textbf{94.1} & 94.2 & \textbf{94.1} \\
    \bottomrule
    \end{tabular*}
    \label{tab:wtw}
\vspace{-1.8em}
\end{table}

\subsection{Ablation Studies}
We conduct multiple ablation experiments on the WTW dataset to verify the effectiveness of different module designs.

\smallskip\noindent\textbf{The effectiveness of module design.}
We conduct ablation experiments on the module design of GridFormer, specifically the query selection module and two-stream decoder design. By removing the query selection module, we use the learnable reference point embedding to initialize the decoder reference point input, which is not relevant to the input table images. GridFormer can bring a +7.8\% improvement (from 86.3\% to 94.1\%) by adopting the query selection module, which shows that better recognition accuracy can be achieved with the aid of accurate reference points initialization.
As mentioned in section ~\ref{subsub:grid_pred}, we also list the result of using a one-stream decoder to predict valid row numbers, valid column numbers, vertexes position, and the valid right/down edge. As shown in Table~\ref{tab:ablation1}, compared with one-stream decoder, the design of two-stream decoder can bring +23.9\% improvement (from 70.2\% to 94.1\%), which demonstrates the effectiveness of two-stream decoder. The two-stream decoder can not only ease the difficulty of regressing vertexes by decoupling the x-axis and y-axis value, but also extract robust embedding for predicting right/down edge.

\begin{table}[t]
\small
\setlength{\tabcolsep}{1mm} 
\setlength{\abovecaptionskip}{0em}
\caption{Ablation studies of module design on WTW.}
\label{tab:ablation1}
\begin{center}
\begin{tabular}{cc|ccc}
    \hline\noalign{\smallskip}
     Query selection & Two stream decoder &  Prec.(\%)& Rec.(\%) & F1-score(\%) \\
     \hline
                & \checkmark & 86.3 & 86.3 & 86.3 \\
     \checkmark & & 70.3 & 70.1 & 70.2 \\
     \checkmark & \checkmark & \textbf{94.1} & \textbf{94.2} & \textbf{94.1} \\
    \noalign{\smallskip}
    \hline
\end{tabular}
\end{center}
\vspace{-1em}
\end{table}

\begin{table}
\small
\setlength{\tabcolsep}{1mm} 
\setlength{\abovecaptionskip}{0em}
\caption{Ablation studies of edge classification on WTW.}
\label{tab:ablation2}
\begin{center}
\begin{tabular}{cc|ccc}
    \hline\noalign{\smallskip}
     query embedding & visual feat &  Prec.(\%)& Rec.(\%) & F1-score(\%) \\
     \hline
     \checkmark & & 63.4 & 61.1 & 62.2 \\
                & \checkmark & 89.0 & 88.5 & 88.7 \\
     \checkmark & \checkmark & \textbf{94.1} & \textbf{94.2} & \textbf{94.1} \\
    \noalign{\smallskip}
    \hline
\end{tabular}
\end{center}
\end{table}

\smallskip\noindent\textbf{The effectiveness of edge classification features.}
We conduct the following ablation studies to further examine the effectiveness of features for edge classification. Based on the row, column query predictions and the matched indices from the Hungarian algorithm, we re-sort the row query predictions on the y-axis and the column queries on the x-axis. Combining the coordinates output from query prediction, we form the $M \times N$ grid and use feature on each vertexes to predict the down/right edge. When only using query embedding for classification, we repeat the row embedding for vertexes on the same row to predict the down-link edge. Similarly, the column embedding is repeated on the same column to predict the right-link edge. When only using visual features for classification, we directly use position prediction to sample visual features on CNN visual backbone.
As shown in Table~\ref{tab:ablation2}, when concatenating the features of query embedding and visual features on vertexes for edge classification, the model achieves the best performance of 94.1\%. This feature combination utilizes both global and local features, with the query embedding conducting self-attention globally and the visual features focusing on local features. This combination helps the model achieve higher classification accuracy.

\section{Conclusion}
In this paper, we present GridFormer, a table structure recognition approach based on the insight that all tables can be represented by a grid. Our method involves two main contributions: (1) a novel, flexible grid representation for tables, and (2) a DETR-style network for predicting tables' geometric information. GridFormer is a straightforward and powerful approach that can handle complex tables effectively. We conducted extensive experiments on various types of tables, including wireless, oriented, multi-merge-cell, and distorted tables, and our method achieves impressive performance in all cases.

\bibliographystyle{ACM-Reference-Format}
\balance
\bibliography{sample-sigconf}

\newpage
\appendix

\section{Implementation}
\subsection{Label generation}
\subsubsection{Labels of reference point} 
\begin{figure}[h]
  \includegraphics[width=0.45\textwidth]{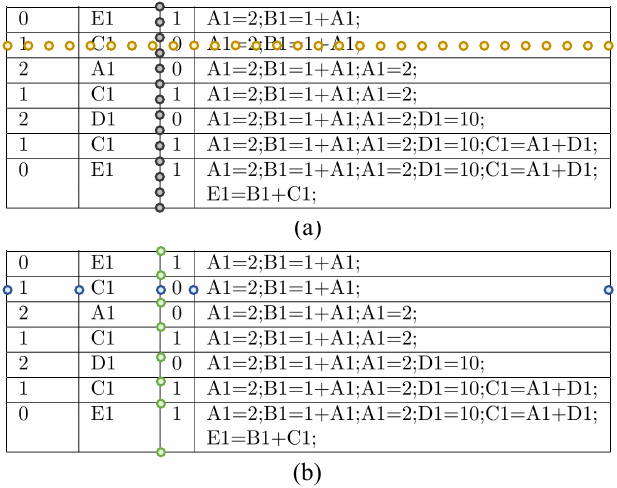} 
    \caption{Visualization of proposal points (a) and the training labels of reference points for rows and columns (b).}
    \label{fig:vis_label}
\end{figure}

The reference point supervision is added to the query selection module and the grid prediction module.
In the query selection module, the row/column proposals output the positive probability and the regression results.
As shown in Figure~\ref{fig:vis_label}(a), the row and column proposal points are overlaid on the input image. These proposals are generated based on the encoded feature map, where the row proposals are evenly located at position ($\tau_{1}$, $y_{i}$), and the column proposals are evenly located at position ($x_{i}$, $\tau_{2}$).
Here, $l$ is the level indice, $\tau_{1}=W/4$, $\tau_{2}=H/4$, $x_{i}\in[0, W^{l}]$ with step of 1, $y_{i}\in[0, H^{l}]$ with step of 1, and $H^{l}, W^{l}$ are the dimension of the encoded feature map. Different from ~\cite{deformable_detr}, we generate proposal points at fixed position rather than the whole feature map, which 
can reduce the redundant proposals and alleviate the matching difficulty. In figure~\ref{fig:vis_label}(b), we visualize the label of reference points on rows and columns. The row reference points have fixed value at x-axis, and the y-axis value is the y-mean value of points on the corresponding same row. Similarly, the column reference points have fixed value at y-axis, and the x-axis value is the x-mean value of points on the corresponding same column. In the grid prediction module, the learned row queries and column queries also regress the reference points, with the training labels being the same as those in the query selection module.

\subsubsection{Labels of vertexes and edges} 
\begin{figure}[h]
  \includegraphics[width=0.45\textwidth]{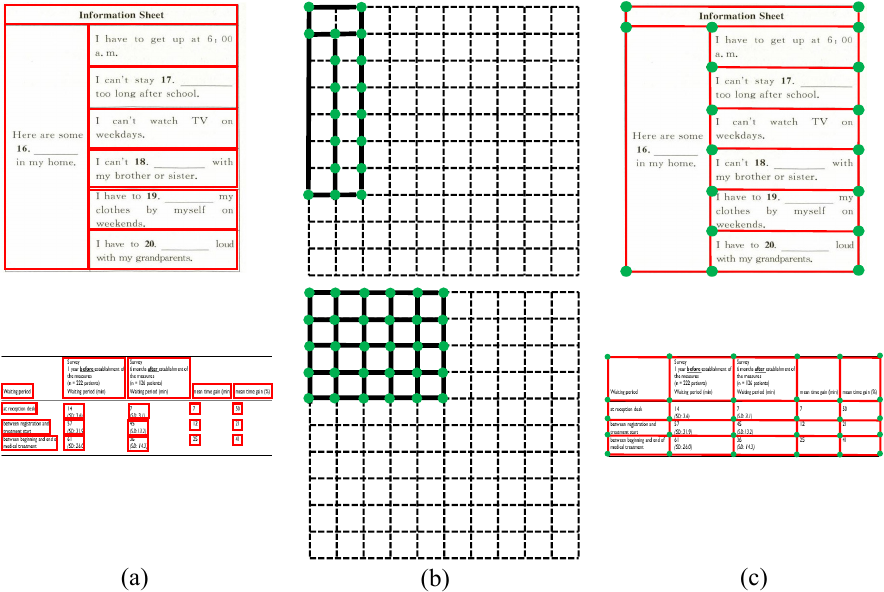} 
    \caption{Visualization of labels. (a) is the original cell bounding box annotation, (b) is the corresponding grid representation, (c) is the table reconstruction output overlaid on the input images.}
    \label{fig:vis_vertex}
\end{figure}

As shown in Figure~\ref{fig:vis_vertex}, we list the visualization results of the original bounding box annotation, training labels, and the table reconstruction visualization. Vertexes on the padding grid that share the same logical indices (same row and column indices) as those on the table will be regarded as positive vertexes, and the corresponding physical coordinates will be inherited. Edges belonging to the boundaries of the cells will be defined as positive edges. Based on the grid representation information, we can reconstruct the table correctly, which can be viewed in Figure~\ref{fig:vis_vertex} (c).

Below, we provide an explanation on how to obtain the vertexes and edges on the table. Tables are annotated with HTML structures and cell bounding boxes. We can obtain the start and end indices of cells by analyzing the HTML tags. The bounding boxes in the table have two types of annotation formats: one based on the cell-level~\cite{cycle_center_net, tal2021}, as shown in the first row of Figure~\ref{fig:vis_vertex} (a), and the other based on the text-line level within the cells~\cite{pubtabnet, GTE, graphTSR}, as shown in the second row of Figure~\ref{fig:vis_vertex} (a). For bounding boxes annotated with the cell-level format, we can directly obtain the vertexes and edges on the table. However, for bounding boxes annotated with the text-line level format, additional steps are required. Firstly, we extend the x-axis coordinates of each column to the outer boundary until they align with the same x-value for that column, and extend the y-axis coordinates of each row to the outer boundary until they align with the same y-value for that row, thus updating the new cell coordinates. The average of the y-axis coordinates of two adjacent rows is the row separator, and the average of the x-axis coordinates of two adjacent columns is the column separator. Finally, we build vertexes and edges based on the updated cell coordinates.

\section{Additional ablation study}

\subsection{The ablation studies on varying M and N}
We conduct ablation experiments on different settings of row query number M and column query number N to verify the model robustness. As shown in Table~\ref{tab:tal_num}, we conduct experiments on TAL dataset with three settings. In the reported result on TAL dataset, the values of M and N are set to 50 and 50, respectively. The results of TEDS-struct and F-score show that the values of M and N have less impact on the performance of structure recognition and cell localization. A larger number of queries can bring a high recall while incurring additional computational costs. Therefore, in our implementation, we set the number of M and N to the largest numbers of rows and columns in the corresponding dataset for computational efficiency.

\begin{table}[h]
\caption{ Ablation studies of varying M and N on TAL dataset.}  
\centering
\setlength{\tabcolsep}{12pt}
\begin{tabular}{lcc}
\hline
 \multirow{2}*{Number Setting} & \multicolumn{2}{c}{TAL}\\
 \cline{2-3}
 & TEDS-Struct &  F-Score  \\
\hline
 M=50, N=50    & 99.4  & \textbf{98.9} \\ 
 M=70, N=70    & \textbf{99.5}  & \textbf{98.9} \\ 
 M=100, N=100  & 99.4  & \textbf{98.9} \\ 
\hline
\end{tabular}
\label{tab:tal_num}
\end{table}

\subsection{The ablation studies on varying $\tau_{1}$ and $\tau_{2}$}
In the table reconstruction stage, we use two thresh $\tau_{1}$ and $\tau_{2}$ to filter positive row/column predictions and positive edges. In this stage, we conduct ablation experiments to discuss how the results are changing by varying these two thresholds. In the reported results, we set $\tau_{1}$ and $\tau_{2}$ to 0.5 and 0.4, respectively, for all evaluated datasets. The reported TEDS-struct and F-score results on the TAL dataset are 99.4\% and 98.9\%, respectively.  As shown in Table~\ref{tab:tal_thresh}, we conduct multiple combination experiments while varying the settings of $\tau_{1}$ within the range of [0.4, 0.6] and the settings of $\tau_{2}$ within the range of [0.3, 0.5]. There is almost no difference on TEDS-struct and F-score results with different settings, which has shown the table reconstruction step is not sensitive to the hyper-parameter settings.

\begin{table}[h]
	\centering
	\caption{Ablation studies of $\tau{1}$ and $\tau_{2}$ on TAL dataset.}
	\begin{tabular*}{\hsize}{@{\extracolsep{\fill}}lcccc}
		\toprule
		     & \makecell[l]{$\tau_{1}$=0.4 \\ $\tau_{2}$=0.4} & \makecell[l]{$\tau_{1}$=0.45 \\ $\tau_{2}$=0.4} & \makecell[l]{$\tau_{1}$=0.55 \\ $\tau_{2}$=0.4} &  \makecell[l]{$\tau_{1}$=0.6 \\ $\tau_{2}$=0.4}  \\
		\midrule
		TEDS-struct       & \textbf{99.40} & \textbf{99.40} & 99.39  & 99.38          \\
		F-Score           & \textbf{98.95} & 98.93 & 98.93  & 98.92          \\
		\bottomrule
		     & \makecell[l]{$\tau_{1}$=0.5 \\ $\tau_{2}$=0.3} & \makecell[l]{$\tau_{1}$=0.5 \\ $\tau_{2}$=0.35} & \makecell[l]{$\tau_{1}$=0.5 \\ $\tau_{2}$=0.45} &  \makecell[l]{$\tau_{1}$=0.5 \\ $\tau_{2}$=0.5}  \\
            \midrule
		TEDS-struct       & 99.39 & 99.39 & \textbf{99.40}  & 99.39          \\
		F-Score           & 98.92 & 98.92 & 98.93  & 98.93          \\
            \bottomrule
	\end{tabular*}
	\label{tab:tal_thresh}
\end{table}

\subsection{Additional qualitative results}
\begin{figure*}[t]
\centering
  \includegraphics[width=0.93\textwidth]{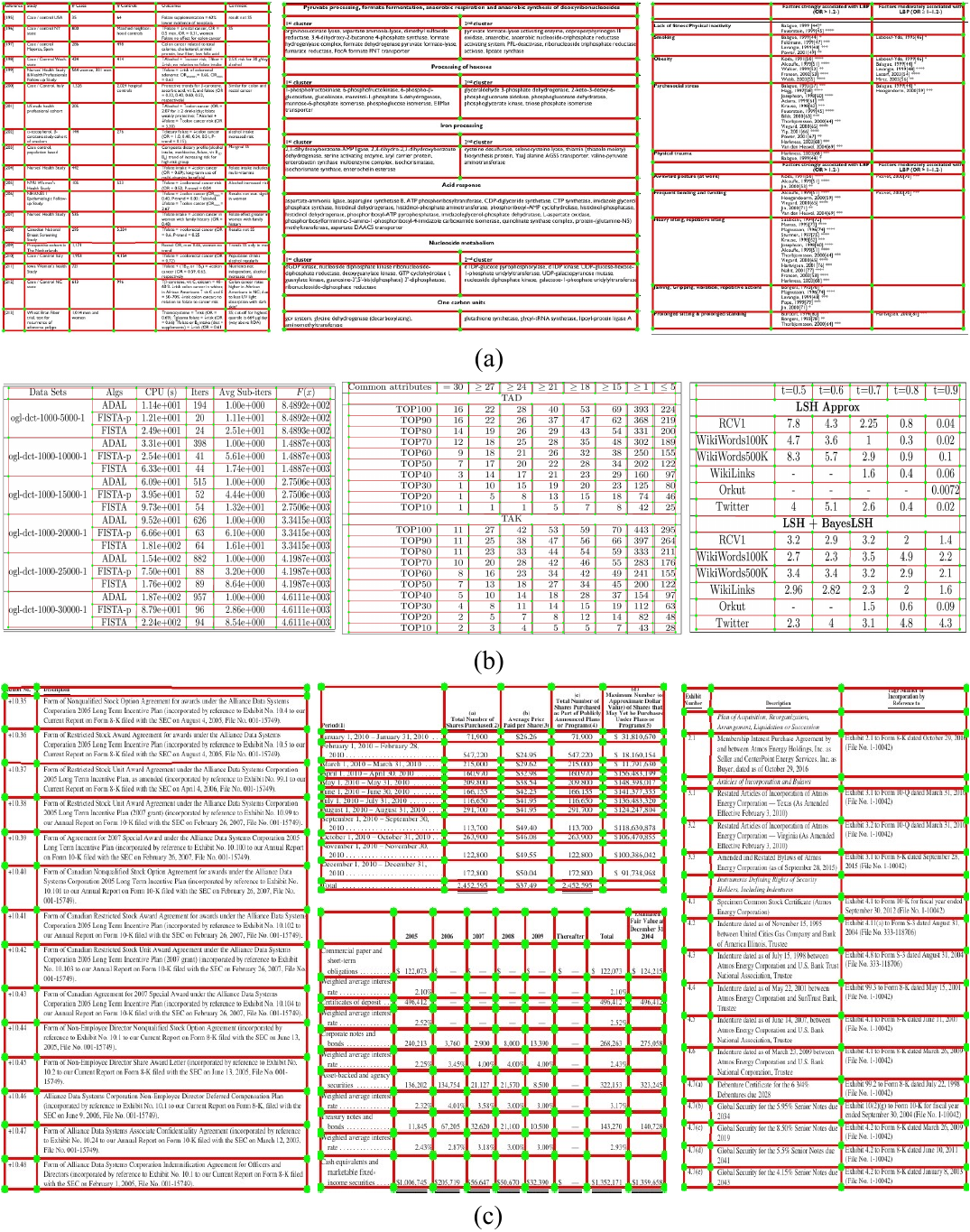} 
    \caption{Visualization results on Regular tables.}
    \label{fig:vis_prediction1}
\end{figure*}

\begin{figure*}[t]
\centering
  \includegraphics[width=0.93\textwidth]{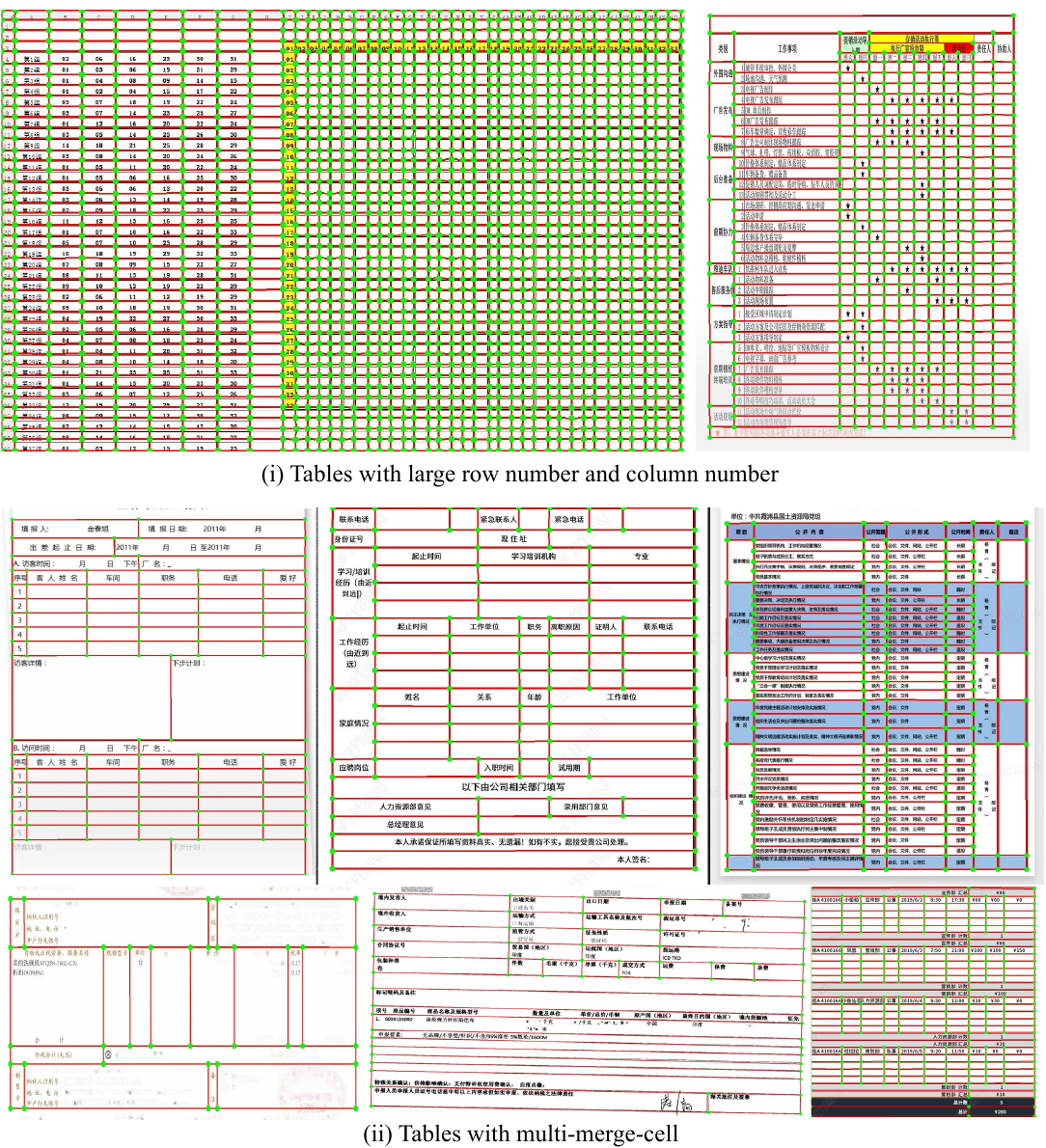} 
    \caption{Visualization results on WTW dataset.}
    \label{fig:vis_prediction2}
\end{figure*}

\begin{figure*}[t]
\centering
  \includegraphics[width=0.85\textwidth]{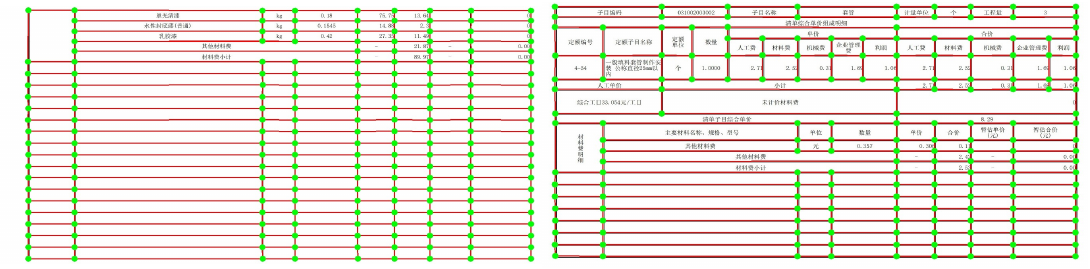} 
    \caption{Visualization results on TAL dataset.}
    \label{fig:vis_prediction3}
\end{figure*}

\begin{figure*}[htb]
\centering
  \includegraphics[width=0.85\textwidth]{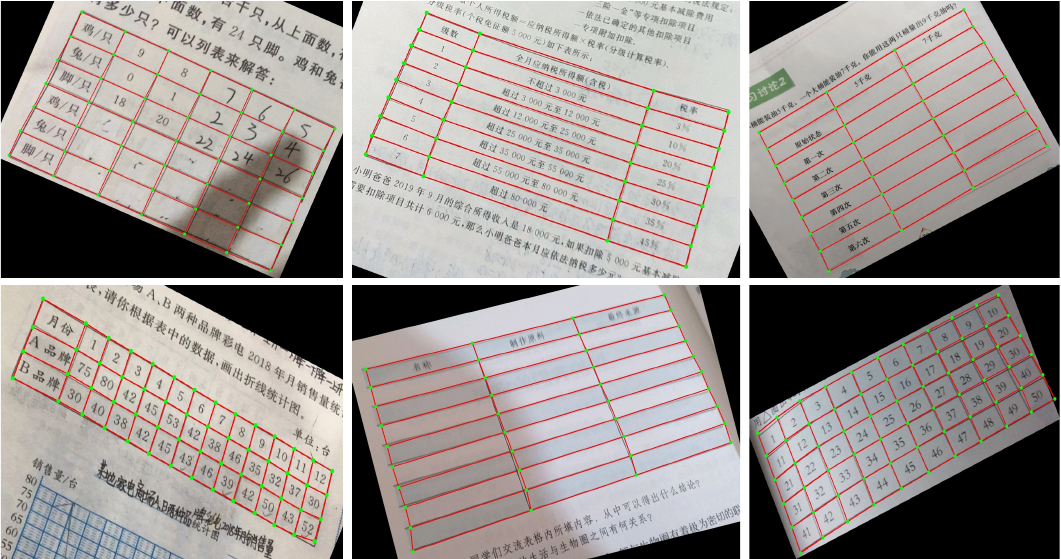} 
    \caption{Visualization results on TAL\_rotated dataset.}
    \label{fig:vis_prediction4}
\end{figure*}

\begin{figure*}[htb]
\centering
  \includegraphics[width=0.85\textwidth]{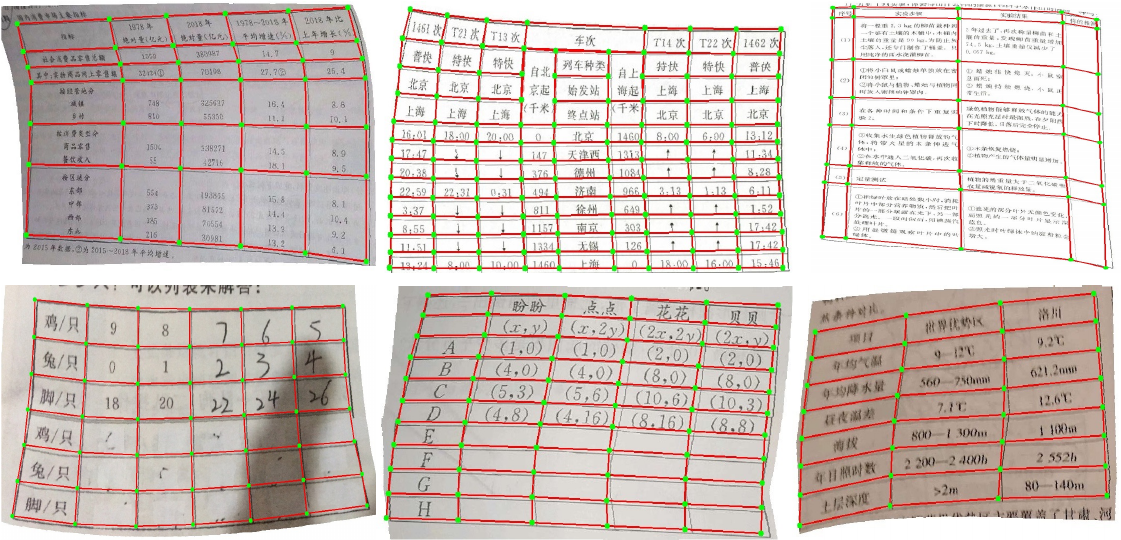} 
    \caption{Visualization results on TAL\_curved dataset.}
    \label{fig:vis_prediction5}
\end{figure*}

In this section, we present additional qualitative results to further demonstrate the effectiveness of our proposed method. As shown in Figure~\ref{fig:vis_prediction1}, we list the visualization results of our methods on tables scanned from documents, including PubTabNet dataset, SciTSR dataset and FinTabNet dataset. These results demonstrate the robustness of our method in recognizing cells with multiple text lines, tables with blank cells, and tables with merged cells. As shown in Figure~\ref{fig:vis_prediction2} and Figure~\ref{fig:vis_prediction3}, based on the design of row query predictions and column query predictions, our method can accurately recognize dense cells, even in cases where the table in (i) contains close to 1600 cells (40 rows x 40 columns). The visualization results on complex tables, where tables have multi-merge-cells and dense cells, demonstrate the potential of the proposed method. Additionally, we demonstrate our method's ability to recognize tables with rotation and distortion in Figure~\ref{fig:vis_prediction4} and Figure~\ref{fig:vis_prediction5}.

Thanks to the flexible grid representation and the decoupled predictions design, we can effectively recognize tables with different structures and shapes accurately. Compared with most existing works, our method can cover many scenarios, while most existing methods are not robust to these challenging scenarios.

\end{document}